\documentclass[10pt,twocolumn,letterpaper]{article}

\usepackage{cvpr}              
\usepackage[accsupp]{axessibility}

\usepackage{multirow}
\usepackage{booktabs}
\usepackage{comment}

\definecolor{cvprblue}{rgb}{0.21,0.49,0.74}
\usepackage[pagebackref,breaklinks,colorlinks,allcolors=cvprblue]{hyperref}

\def\paperID{18} 
\def\confName{CVPR}
\def\confYear{2026}

\title{DiffusionPrint: Learning Generative Fingerprints \\for Diffusion-Based Inpainting Localization}

\author{Paschalis Giakoumoglou \qquad Symeon Papadopoulos\\
Information Technologies Institute, CERTH\\
{\tt\small \{giakoupg, papadop\}@iti.gr}\\
}

\begin{document}
\maketitle
\begin{abstract}
Modern diffusion-based inpainting models pose significant challenges for image forgery localization (IFL), as their full regeneration pipelines reconstruct the entire image via a latent decoder, disrupting the camera-level noise patterns that existing forensic methods rely on. We propose DiffusionPrint, a patch-level contrastive learning framework that learns a forensic signal robust to the spectral distortions introduced by latent decoding. It exploits the fact that inpainted regions generated by the same model share a consistent generative fingerprint, using this as a self-supervisory signal. DiffusionPrint trains a convolutional backbone via a MoCo-style objective with cross-category hard negative mining and a generator-aware classification head, producing a forensic feature map that serves as a highly discriminative secondary modality in fusion-based IFL frameworks. Integrated into TruFor, MMFusion, and a lightweight fusion baseline, DiffusionPrint consistently improves localization across multiple generative models, with gains of up to +28\% on mask types unseen during fine-tuning and confirmed generalization to unseen generative architectures. Code is available at \url{https://github.com/mever-team/diffusionprint}
\end{abstract}

\section{Introduction}
\label{sec:intro}

\begin{figure}[t]
  \centering
  \includegraphics[width=\linewidth]{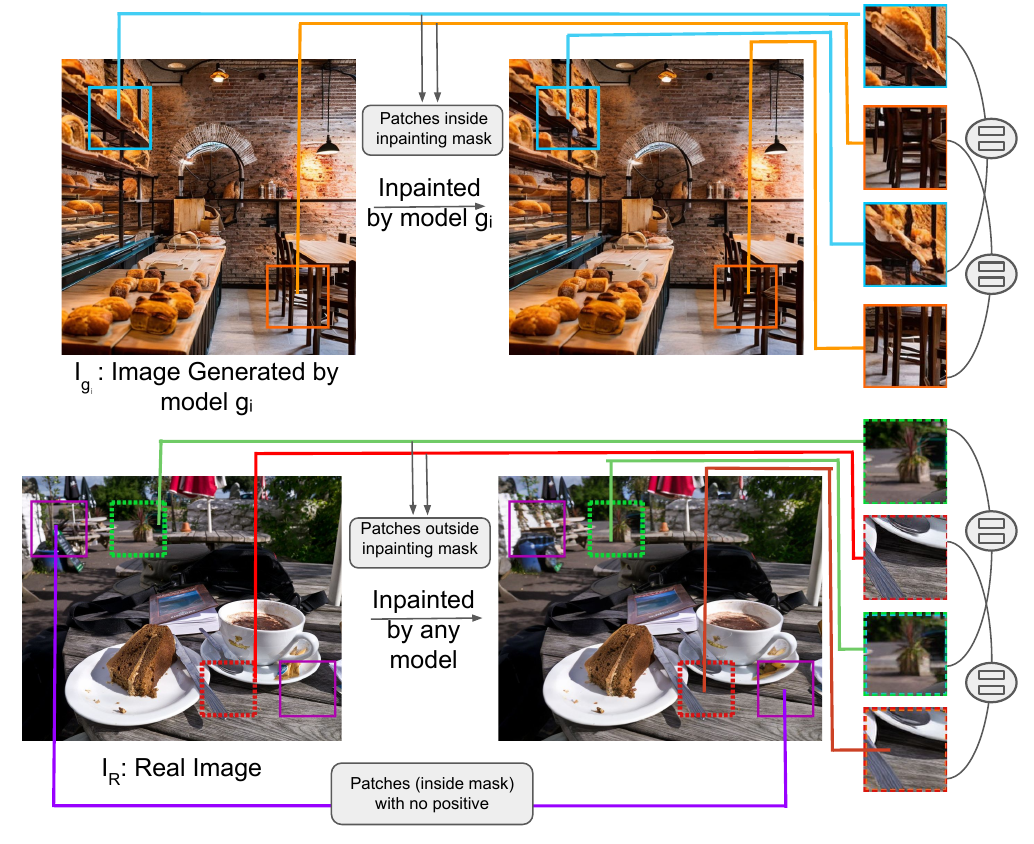}
  \caption{Positive pair construction for the two image categories in our contrastive framework. An image generated by a generator $g_i$ is inpainted using the corresponding inpainting variant of $g_i$. Natural images are also inpainted. For generated images, we extract the inpainted patches and pair them as positives with patches from the exact same spatial position in the source image. For natural images, positive pairs are formed by sampling patches outside the inpainting mask in both the source and the inpainted image.}
  \label{fig:positive_construction}
\end{figure}

Digital image manipulation has undergone a fundamental shift with the rise of generative AI. While convincing forgeries once required expertise in tools like Adobe Photoshop or specialized deepfake pipelines, modern text-guided inpainting models such as Stable Diffusion~\cite{rombach2022sd}, FLUX~\cite{flux2024}, and Adobe Firefly~\cite{adobe2023firefly} allow non-experts to produce photorealistic edits through simple textual prompts. The resulting images are increasingly indistinguishable from authentic content~\cite{hessen2024advancesinaigen}, posing serious risks for disinformation, fraud, and the fabrication of visual evidence~\cite{verdoliva2020media}.

Among the manipulation strategies enabled by these models, text-guided inpainting is particularly challenging for forensic analysis. In traditional splicing workflows, only the pixels within the manipulated region are replaced, leaving the rest of the image untouched and preserving low-level forensic traces such as sensor noise. However, modern latent diffusion-based inpainting operates in a fully-regenerative manner: generation and blending occur entirely within a compressed latent space~\cite{avrahami2023blendedlatentdiffusion, ju2024brushnet}, and the final output is produced by decoding the entire latent representation through a Variational Autoencoder (VAE) decoder. As a result, all pixels are globally reconstructed, even in regions outside the inpainting mask that remain semantically unchanged. This process destroys and renders ineffective the localized forensic traces that traditional image forgery localization (IFL) methods depend on~\cite{mareen2024tgif, giakoumoglou2025sagi}.

Existing forensic methods are designed around the splicing threat model. A dominant family of approaches suppresses image content to reveal low-level acquisition artifacts, treating discontinuities in these signals as evidence of manipulation~\cite{cozzolino2020noiseprint, mahdian2009noise, guillaro2023trufor}. These signals are reliable when authentic regions preserve their original noise statistics, but the global latent reconstruction induces a pervasive spectral shift across all pixels~\cite{nebioglu2026aigeneratedimagedetectorsoverrely}, eliminating the distributional contrast these methods depend on.

We argue that fully-regenerative inpainting leaves a qualitatively different forensic trace that existing pipelines are not designed to capture. An effective forensic signal must be invariant to the global spectral shift imposed by latent reconstruction, while remaining sensitive to whether a region is authentic or generated. To achieve this, we propose \textbf{DiffusionPrint}, a patch-level contrastive learning framework that learns a generative fingerprint suited to this setting. To achieve this, we design a contrastive pairing strategy, illustrated in Figure~\ref{fig:positive_construction}, with two complementary components: (a) for authentic patches, we pair the original content with its latent-reconstructed counterpart, explicitly teaching the encoder to suppress VAE reconstruction artifacts and become invariant to the global spectral shift; (b) for generated patches, we start from a fully AI-generated image, perform inpainting, and pair the newly inpainted patch with its pre-inpainting counterpart at the same spatial location. Since both views originate from the same latent diffusion model, they share the same underlying generative fingerprint~\cite{wang2023dire}, providing the diverse views necessary for robust contrastive learning~\cite{chen2020simclr}. DiffusionPrint trains a convolutional backbone via a MoCo~\cite{he2020moco} style objective with cross-category hard negative mining and a generator-aware classification head, serving as a drop-in replacement for noise-based modalities in fusion-based IFL frameworks. We evaluate on the TGIF~\cite{mareen2024tgif} benchmark across SD~2.1, SDXL, and Flux variants, integrated into three fusion-based IFL frameworks.

Our contributions are as follows:
\begin{itemize}
    \item We propose DiffusionPrint, a patch-level contrastive learning framework with asymmetric positive pair construction and a generator-aware classification head, designed to learn a forensic signal that is invariant to latent reconstruction artifacts on authentic regions while capturing the generative fingerprint of inpainted content.
    \item We introduce a MoCo-style training procedure with a cross-category hard negative queue and top-$k$ mining, enabling effective contrastive learning across generative model fingerprints. 
    \item We evaluate DiffusionPrint integrated into three fusion-based IFL frameworks on the TGIF benchmark, demonstrating consistent improvements of up to +28\% over Noiseprint++ as a forensic signal, with generalization to unseen generative architectures confirmed.
\end{itemize}

\section{Related Work}

\subsection{Generative Inpainting}
 
Text-guided inpainting has advanced rapidly through diffusion and flow-based architectures~\cite{rombach2022sd, podell2023sdxl, flux2024}, with methods ranging from training-free latent blending~\cite{avrahami2023blendedlatentdiffusion}, to fine-tuned inpainting models~\cite{nichol2021glide, podell2023sdxl, wang2023imagen}, to adapter-based approaches~\cite{ju2024brushnet, manukyan2024hdpainter, zhuang2024powerpaint}. In all cases, the generation process operates entirely in latent space and the final image is obtained by decoding the full latent representation through a VAE decoder, including regions outside the inpainting mask. This global latent reconstruction modifies all pixels, destroying the noise statistics that IFL methods rely on~\cite{mareen2024tgif, giakoumoglou2025sagi} and inducing a pervasive spectral shift that detectors tend to exploit as a shortcut rather than identifying locally synthesized content~\cite{nebioglu2026aigeneratedimagedetectorsoverrely}. Some pipelines apply post-hoc blending to embed the generated region back onto the original image~\cite{ju2024brushnet, zhuang2024powerpaint, yu2023inpaintanything, manukyan2024hdpainter}, producing effectively spliced outputs where existing IFL methods remain applicable~\cite{giakoumoglou2025sagi, mareen2024tgif}; our work targets the harder fully-regenerative setting. Several datasets support forensic research on text-guided inpainting~\cite{guillaro2023trufor, jia2023autosplice, yang2025coinco, sun2024gre, wang2025opensdi}, yet few address the fully-regenerative setting~\cite{giakoumoglou2025sagi, mareen2024tgif, mareen2026tgif2extendedtextguidedinpainting}, and prior work has shown that existing localization models fail on such images without retraining~\cite{giakoumoglou2025sagi, mareen2026tgif2extendedtextguidedinpainting}.

\subsection{Forgery Localization and Forensic Fusion}
 
Early IFL approaches relied on handcrafted signal processing to detect manipulation artifacts, including structural inconsistencies~\cite{wu2008detection, chang2013forgerydetection, liang2015efficientforgerydetection, zhang2018robustforgery}, acquisition traces such as sensor noise and CFA patterns~\cite{ferrara2012cfa, chierchia2011prnu, lukas2006prnu, zhang2023prnu}, and statistical artifacts from processing and compression~\cite{cozzolino2015splicebuster, bianchi2011improved, farid2009jpegghost, iakovidou2018jpeggrid, nikoukhah2021zero}. Recent deep learning methods, both CNN-based~\cite{wu2019mantranet, liu2022pscc, rajini2019image, zhang2016image, dong2022mvss, wu2022iid} and transformer-based~\cite{li2017localization, liu2023tbformer, guillaro2023trufor, triaridis2024mmfusion}, have substantially advanced the state of the art across splicing, copy-move, and conventional inpainting. Another line of work combines multiple forensic signals within fusion frameworks~\cite{zhou2018twostream}. TruFor~\cite{guillaro2023trufor} introduces a dual-encoder Segformer that fuses RGB features with Noiseprint++, a learned camera-model fingerprint, establishing a modular pattern where the forensic signal extractor and the localization architecture are decoupled. MMFusion~\cite{triaridis2024mmfusion} and OMGFuser~\cite{karageorgiou2024fusion} extend this paradigm using more expressive fusion architectures. Despite this progress, existing models have not been optimized for fully-regenerative inpainting and their performance degrades substantially in this setting~\cite{giakoumoglou2025sagi}, where the distributional contrast between authentic and manipulated regions is absent.

\subsection{Contrastive Learning}

Contrastive self-supervised learning has emerged as a powerful paradigm for learning visual representations without manual labels, by training an encoder to attract positive pairs while repelling negatives in embedding space~\cite{giakoumoglou2025reviewdiscriminativeselfsupervisedlearning}. Early influential works such as SimCLR~\cite{chen2020simclr} demonstrated that carefully designed augmentation policies and large numbers of negatives are critical for representation quality, while methods such as BYOL~\cite{grill2020byol} and DINO~\cite{caron2021dino} showed that competitive representations can be learned without explicit negatives through asymmetric architectures and self-distillation. Other approaches relax the rigid contrastive objective by replacing hard positive–negative assignments with soft targets~\cite{zheng2021ressl, giakoumoglou2024rrd}. MoCo~\cite{he2020moco} introduced two key mechanisms, a momentum encoder updated via exponential moving average to produce stable embeddings for negatives and positives, and a memory queue that decouples the number of negatives from the batch size enabling efficient training with massive negative sets. Subsequent work has shown that the quality of negatives is as important as their quantity — hard negative mining, which prioritizes negatives that are most similar to the anchor in embedding space, leads to more discriminative representations and faster convergence~\cite{robinson2020contrastive, kalantidis2020mochi, giakoumoglou2025syncosynthetichardnegatives}.

\section{Methodology}

\begin{figure}[t]
    \centering
    \includegraphics[width=\linewidth]{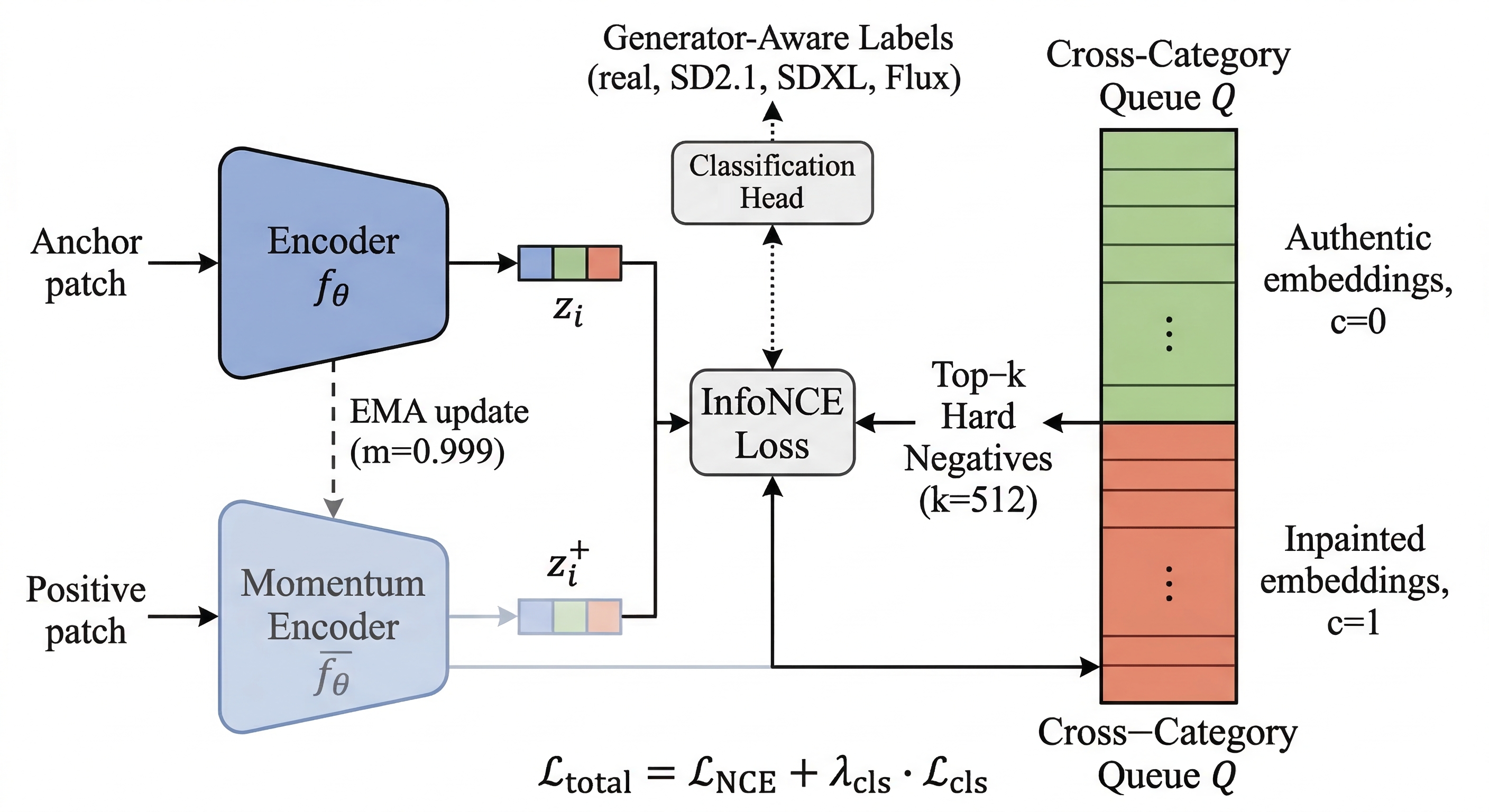}
    \caption{Overview of DiffusionPrint. An anchor patch is encoded by $f_\theta$ and its positive by the momentum encoder $f_{\bar{\theta}}$, updated via exponential moving average. Negatives are drawn from a cross-category queue, where each anchor receives the top-$k$ most similar embeddings from the opposite category. The InfoNCE loss and a generator-aware classification head are jointly optimized.}
    \label{fig:architecture}
\end{figure}

DiffusionPrint is designed as a forensic signal extractor 
in existing fusion-based IFL frameworks, which accept a secondary forensic feature map alongside RGB as input, enabling plug-in replacement for other forensic signals. An overview of the algorithm is shown in Figure~\ref{fig:architecture}.

\subsection{Empirical Motivation}
\label{sec:motivation}

\begin{figure*}[t]
    \centering
    \includegraphics[width=0.33\linewidth]{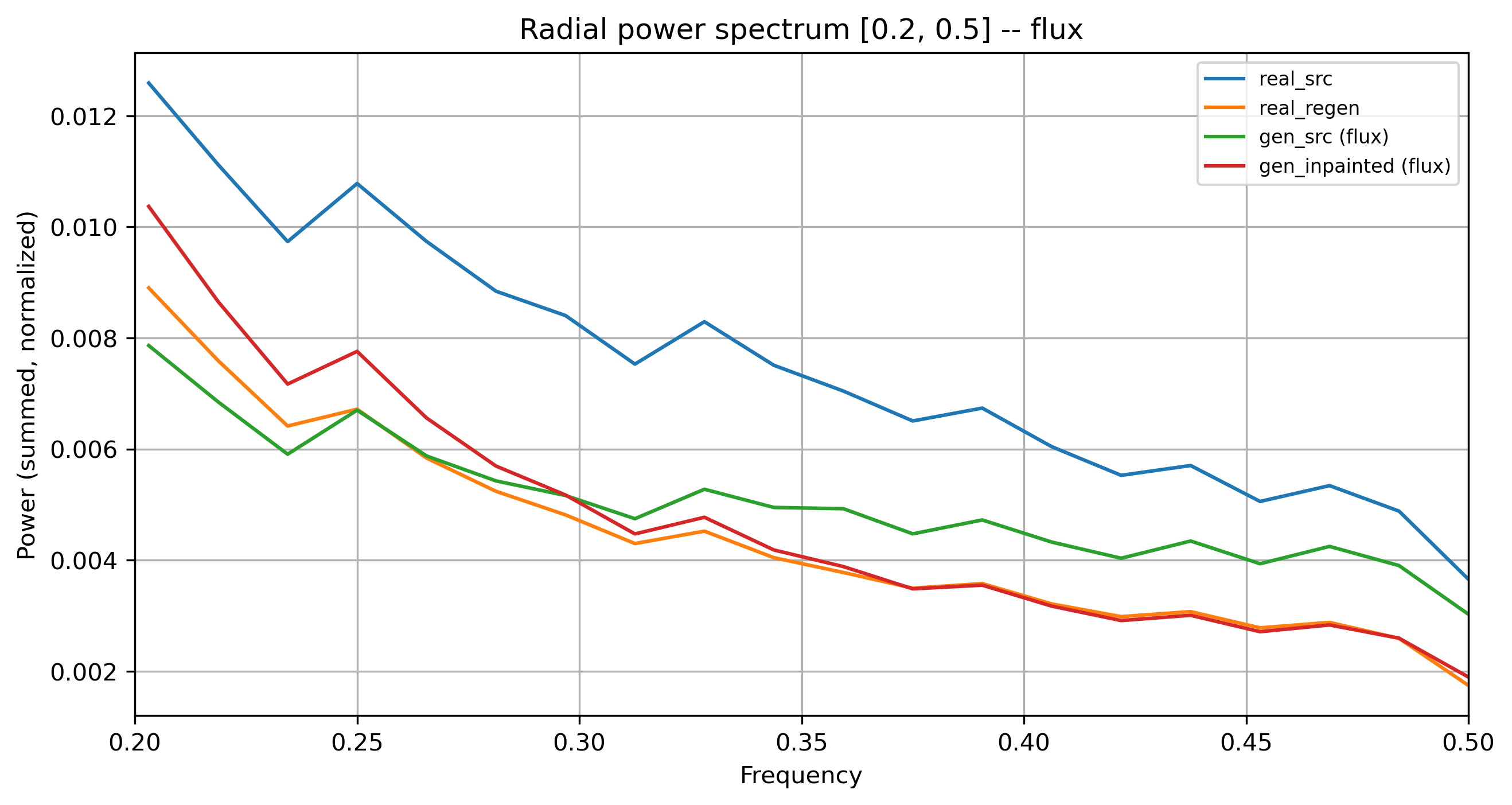}
    \includegraphics[width=0.33\linewidth]{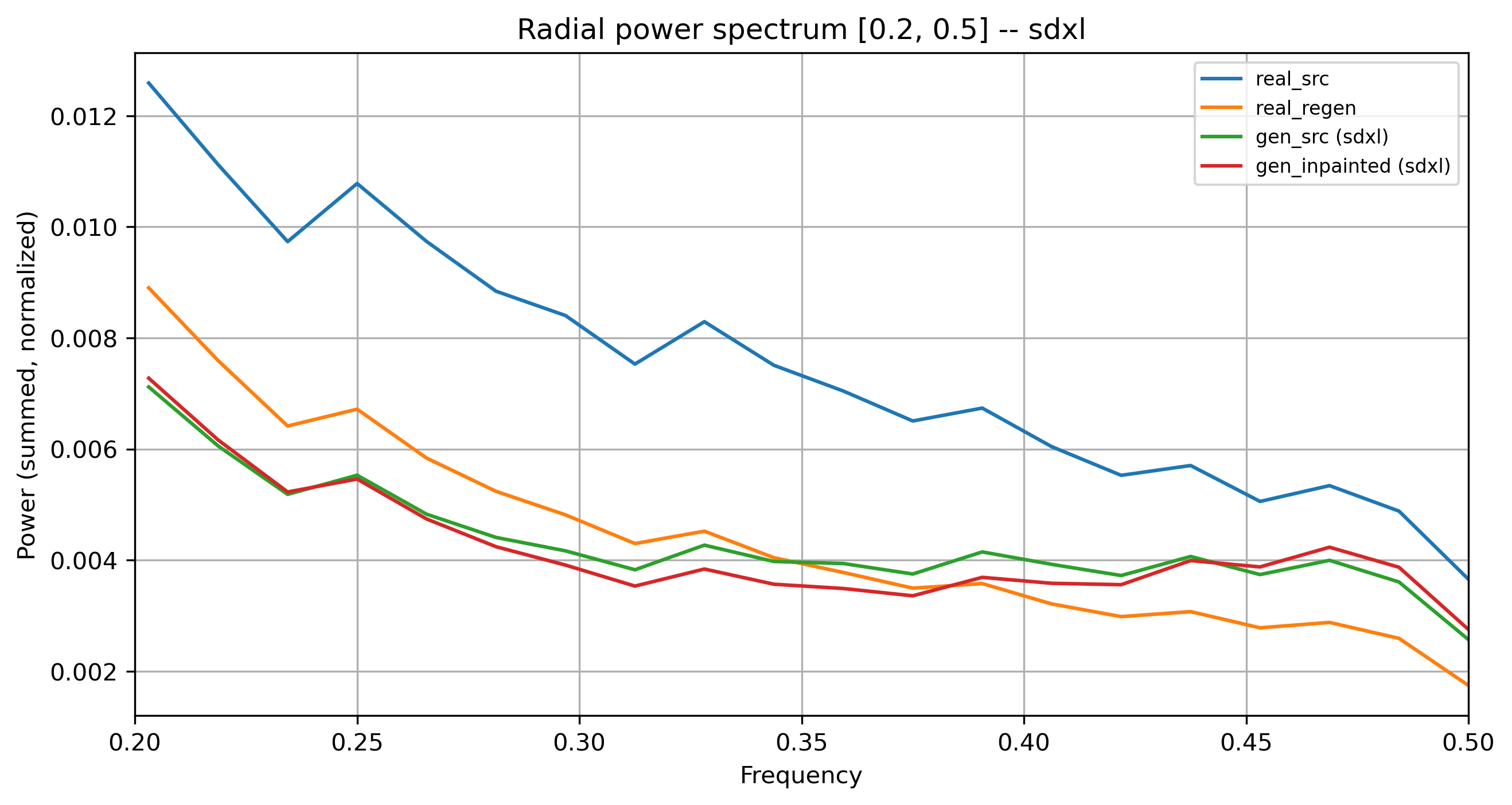}
    \includegraphics[width=0.33\linewidth]{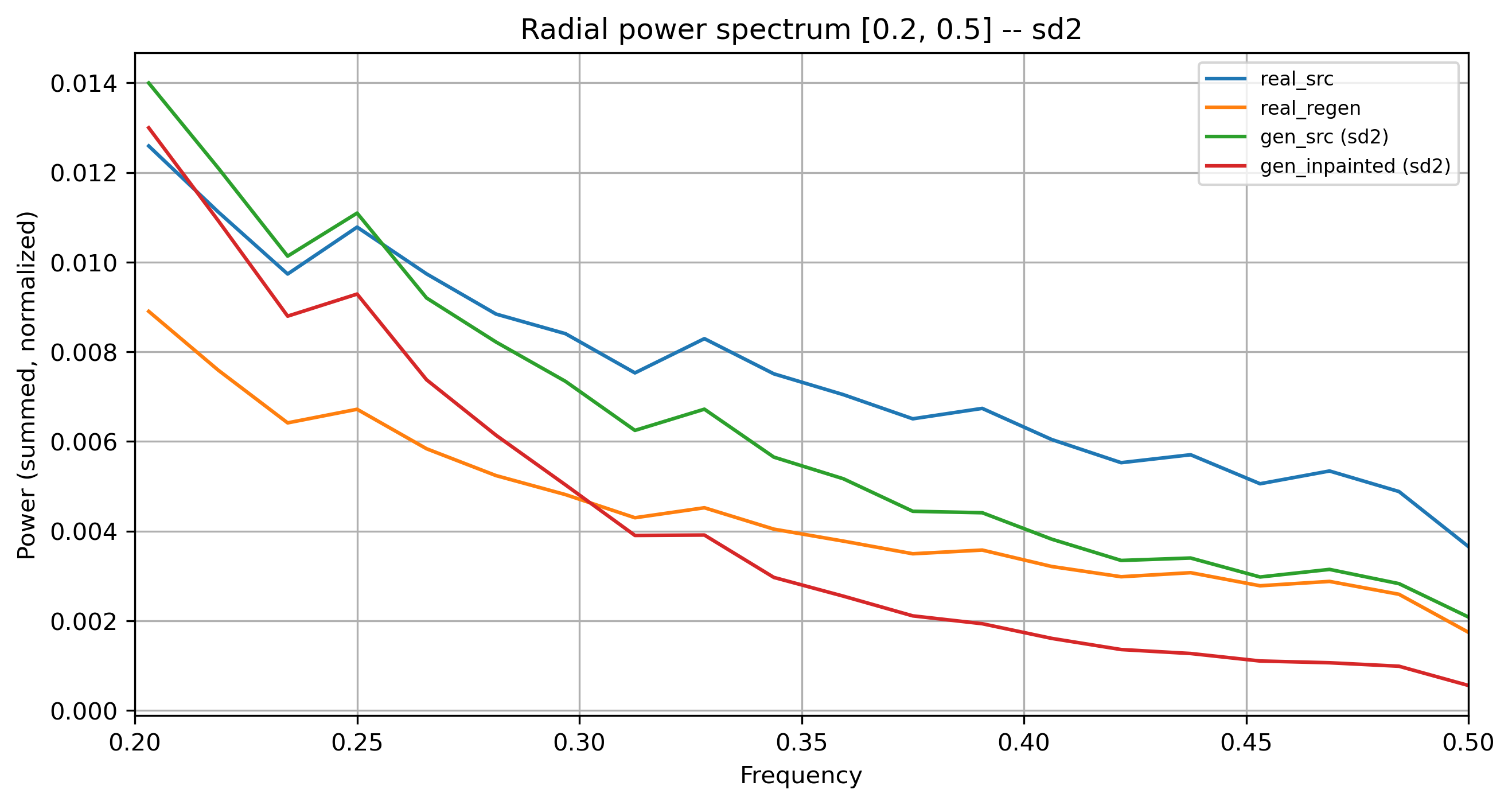}
    \caption{Radial power spectra in the mid-to-high frequency band $[0.2, 0.5]$ for Flux (left), SDXL (center), and SD~2.1 (right). Four patch types are shown: original real (\textcolor[rgb]{0.12,0.47,0.71}{real\_src}), reconstructed real (\textcolor[rgb]{1.0,0.55,0.0}{real\_regen}), original generated (\textcolor[rgb]{0.17,0.63,0.17}{gen\_src}), and inpainted generated (\textcolor[rgb]{0.84,0.15,0.16}{gen\_inpainted}) patches. Across all three models, latent reconstruction attenuates high-frequency content in real patches, shifting their spectral profile toward that of generated content.}
    \label{fig:spectral}
\end{figure*}

Before describing the proposed framework, we present a preliminary spectral analysis that directly motivates our design choices. Corvi~et~al.~\cite{corvi2023intriguingproperties} showed that fully generated synthetic images deviate from the $1/f^\alpha$ power law of natural images in the mid-to-high radial frequency range $[0.2, 0.5]$, and that this deviation provides a useful signal for detecting synthetic content. However, diffusion-based inpainting introduces a complication that does not arise in full-image generation since the entire image, not only the inpainted region, goes through the latent reconstruction process. This means that even authentic regions outside the inpainting mask undergo lossy reconstruction, altering their frequency statistics. 

Figure~\ref{fig:spectral} examines this effect directly. We plot radial power spectra for four patch types (original real, latent-reconstructed real, original generated, and inpainted generated) across three inpainting models. Two patterns are consistent across all models. First, latent reconstruction substantially attenuates the high-frequency content of real patches, shifting their spectral profile toward that of generated content. Second, original generated patches and their inpainted counterparts are already spectrally aligned, as both have passed through the same generative process. This means that any signal relying on absolute frequency statistics will be confounded by the latent reconstruction, which shifts authentic regions into the same spectral range as generated content.

\subsection{Problem Formulation}

Let $\mathcal{I}$ denote a set of images that have been processed by a fully-regenerative inpainting pipeline, where each image $I \in \mathcal{I}$ passes through a generative model regardless of whether a given region was semantically edited. We partition image patches into two categories: inpainted patches $\mathcal{P}^{+}$, which are the direct target of the generative process, and authentic patches $\mathcal{P}^{-}$, which lie outside the inpainting mask but have been reconstructed by the generative model. Our goal is to learn an encoder $f_\theta : \mathbb{R}^{H \times W \times 3} \rightarrow \mathbb{R}^{d}$ that maps patches to a $d$-dimensional embedding space such that inpainted and authentic patches are well-separated, while remaining invariant to incidental latent reconstruction artifacts shared by both categories.

\subsection{Backbone Architecture}
\label{sec:backbone}

DiffusionPrint requires a backbone that preserves the spatial dimensions of its input, producing a dense feature map that can be directly fused with RGB features in downstream localization frameworks. Any fully convolutional architecture that maintains spatial resolution satisfies this requirement. The backbone produces a feature map of shape $(B, C, H, W)$, which during contrastive pretraining is reduced to a $(B, C)$ vector via Global Average Pooling (GAP) and a two-layer MLP projector $g_\phi : \mathbb{R}^C \rightarrow \mathbb{R}^{128}$ with hidden dimension 128, yielding the final patch embedding. Since GAP discards spatial layout, geometric augmentations such as random horizontal and vertical flips are applied during training without affecting the extracted features, effectively providing additional views of the same patch at no cost. At inference, the projector is discarded and the full $(B, C, H, W)$ feature map is passed to the downstream fusion framework, preserving spatial localization through the backbone's dense outputs.

\subsection{Positive Pair Construction}
\label{sec:pairs}

The definition of positive pairs is the core supervisory signal in DiffusionPrint and must be designed to encode reconstruction fidelity asymmetry without introducing conflicting supervision from latent reconstruction artifacts shared across categories.

\paragraph{Positive Pair Construction.} We construct our training data from source images comprising authentic (real) images and images generated by $K$ distinct latent diffusion models, denoted as $g_i$ for $i \in \{1, \dots, K\}$. From each image, we define an $M \times N$ grid to extract spatially aligned patches of size $P \times P$ before and after the inpainting process. We define two types of positive pairs to enforce the complementary invariances required for our forensic signal:

\begin{itemize}
    \item \emph{Generated pairs (inside-mask).} For an image originally generated by $g_i$, we apply inpainting using exclusively the corresponding inpainting variant of $g_i$. We extract our $M \times N$ patches from \textit{inside} the masked region. For each position $(i, j)$ in the grid, the original generated patch $p^{\text{orig}}_{ij}$ and its newly synthesized counterpart $p^{\text{inp}}_{ij}$ form a positive pair $(p^{\text{orig}}_{ij}, p^{\text{inp}}_{ij}) \in \mathcal{P}^{+}$. Because both views originate from the same generative process, this pairing encourages the encoder to isolate the shared generative fingerprint.

    \item \emph{Authentic pairs (outside-mask).} For real images, we inpaint the image $K$ times—once using each of the $K$ generators. We extract the $M \times N$ grid of patches from regions that \textit{do not overlap} with the applied inpainting mask. Consequently, each pristine real patch $p^{\text{real}}_{ij}$ has $K$ latent-reconstructed counterparts (one from each generator's VAE). All pairs within this set of $K+1$ views are treated as positives. This forces the encoder to learn representations that are strictly invariant to latent reconstruction artifacts, capturing the absence of a generative fingerprint consistently across all models.

\end{itemize}

\paragraph{Hard Negatives.} A special case arises from patches at inside-mask positions in real images: these patches have been generated from scratch during inpainting. Treating the generated patches and the respective patches at the same position as positives would undermine the forensic signal, as we do not want the model to produce similar embeddings for patches that originated from real content and patches that were generated at the same position, since this would directly conflict with the goal of localizing generated regions. These patches are therefore excluded from positive pair construction and placed into a dedicated pool alongside other patches for use as negatives.

To guide the model toward the forensically relevant decision boundary, we define hard negatives through cross-category mining from this pool. For each anchor, we mine the most similar patches from the opposite category: if the anchor is a generated patch, we retrieve the most similar real patches from the pool, and vice versa. These represent the most challenging cases for the model, a generated patch that closely resembles real content, or a real patch that closely resembles generated content. Focusing the contrastive signal on these pairs pushes the model to capture the  forensic traces that distinguish the two categories even when their visual appearance is similar.

\subsection{MoCo-Style Contrastive Training}
\label{sec:moco}

We adopt a MoCo-style~\cite{he2020moco} training objective with a momentum encoder to stabilize training with a large number of negatives without requiring large batch sizes.

\paragraph{Momentum encoder.} The encoder $f_\theta$ is paired with a momentum encoder $f_{\bar{\theta}}$, whose parameters are updated via exponential moving average: $\bar{\theta} \leftarrow m\bar{\theta} + (1-m)\theta$, with momentum $m = 0.999$. All positive patch embeddings and queue entries are computed by the momentum encoder, while the anchor is encoded by $f_\theta$.

\paragraph{Cross-category queue.} We maintain a queue $\mathcal{Q}$ of patch embeddings tagged by category $c \in \{0, 1\}$, where $c=0$ denotes authentic and $c=1$ denotes inpainted. For each anchor, negatives are drawn exclusively from the opposite category: inpainted anchors receive authentic negatives and vice versa. This cross-category hard negative mining focuses the contrastive signal on the forensically relevant distinction between inpainted and authentic patches, rather than on incidental content differences within a category.

\paragraph{Top-$k$ hard negative mining.}
From the cross-category queue, we select the top-$k$ most similar negatives per anchor using cosine similarity, focusing gradients on the hardest cases. We use $k=512$  in our primary configuration.

\paragraph{InfoNCE loss.} Let $\mathbf{z}_i = g_\phi(f_\theta(p_i))$ denote the projected embedding of anchor patch $p_i$, and let $\mathbf{z}_i^{+}$ denote the embedding of its positive, computed by the momentum encoder. Let $\{\mathbf{n}_1, \ldots, \mathbf{n}_k\}$ denote the top-$k$ hard negatives drawn from the cross-category queue. The InfoNCE loss for anchor $i$ is:

\begin{equation}
    \mathcal{L}_{\text{NCE}}^{(i)} = -\log \frac{
        \exp\!\left(\mathbf{z}_i \cdot \mathbf{z}_i^{+} / \tau\right)
    }{
        \exp\!\left(\mathbf{z}_i \cdot \mathbf{z}_i^{+} / \tau\right)
        + \sum_{j=1}^{k} \exp\!\left(\mathbf{z}_i \cdot \mathbf{n}_j / \tau\right)
    }
    \label{eq:infonce}
\end{equation}
where $\tau = 0.07$ is the temperature. The batch loss is the mean over all anchors with at least one positive.

\subsection{Classification Head and Joint Objective}
\label{sec:cls}

To further strengthen the forensic signal, we optionally attach a linear classification head $h_\psi : \mathbb{R}^{128} \rightarrow \mathbb{R}^{C_{\text{cls}}}$ on top of the projected embedding, where $C_{\text{cls}} \in \{2, 4\}$ for binary real/inpainted classification or 4-class generative model attribution (real, SD~2.1, SDXL, Flux), respectively. The classification loss is standard cross-entropy:

\begin{equation}
    \mathcal{L}_{\text{cls}} = -\sum_{c=1}^{C_{\text{cls}}} y_c \log \hat{y}_c
    \label{eq:cls}
\end{equation}
where $y_c$ is the ground-truth one-hot label and $\hat{y}_c$ is the predicted probability for class $c$. The combined training objective is:

\begin{equation}
    \mathcal{L}_{\text{total}} = \mathcal{L}_{\text{NCE}} + \lambda_{\text{cls}} \cdot \mathcal{L}_{\text{cls}}
    \label{eq:total}
\end{equation}
where $\lambda_{\text{cls}} \geq 0$ controls the contribution of the classification head. Setting $\lambda_{\text{cls}} = 0$ recovers pure contrastive training. The classification head is used only during training; at inference, DiffusionPrint operates as a pure feature extractor, producing the DnCNN feature map $f_\theta(p)$ as the forensic signal passed to downstream fusion frameworks.
\section{Contrastive Pretraining Evaluation}
\label{sec:pretraining}

In this section, we evaluate the quality of the forensic signal learned by DiffusionPrint independently of any downstream localization framework. We assess representation quality via  linear probing on held-out patches from the Dragon dataset \cite{bertazzini2025dragonlargescaledatasetrealistic}, and ablate key design choices, including the pairing strategy, the number of output channels, classification head weight, and top-$k$ hard negative mining.

\subsection{Experimental Setup}
\label{sec:pretraining-setup}

\paragraph{Backbone.} Following Noiseprint++~\cite{guillaro2023trufor}, we use a denoising convolutional neural network (DnCNN)~\cite{zhang2017dncnn} backbone with 17 convolutional layers, each producing 64-channel feature maps, trained from scratch. The network takes a $3$-channel RGB patch as input and outputs a $(B, 64, H, W)$ feature map that preserves the spatial dimensions of the input.

\paragraph{Dataset.} We use Dragon~\cite{bertazzini2025dragonlargescaledatasetrealistic} dataset for training, which includes 10k real images from COCO val2017 and 10k AI-generated images each from SD~2.1, SDXL, and Flux~1. We extract $64 \times 64$ pixel patches using a $4 \times 4$ grid, yielding up to 16 patches per image. This patch size follows the convention established by Noiseprint and Noiseprint++~\cite{cozzolino2020noiseprint, guillaro2023trufor}, and is particularly suited to our contrastive setup since at this resolution, inpainting a patch with the same model is likely to produce visually similar content, keeping the positive pairs semantically consistent and allowing the encoder to focus on the forensic trace rather than content differences. For generated images, only inside-mask patches are used, yielding 320,000 patches per model ($960{,}000$ total), each paired with its source counterpart as a positive. For real images, outside-mask patches contribute up to 4 views each (source plus one reconstruction per inpainting model); the grid is adapted when patch centers fall below 64 pixels apart, reducing the total below the theoretical $640{,}000$ to $550{,}728$. the resulting dataset comprises 2,150,728 patches in total, of which 1,510,728 are used for contrastive training and 640,000 serve exclusively as hard negatives.

\paragraph{Evaluation protocol.} We evaluate learned representations via linear probing, as it is widely used in contrastive learning setups, training a logistic regression classifier on frozen embeddings following the standard protocol~\cite{he2020moco, chen2020simclr, caron2021dino, grill2020byol}. The evaluation set is constructed independently of the Dragon training split and consists of 180,000 patches sampled from all 25 generative models present in Dragon, alongside 90,000 latent-reconstructed COCO patches and 90,000 unaltered RAISE~\cite{dang2015raise} patches as sources of real content. We apply a 70/30 train/test split and report overall accuracy and in-domain accuracy on the three models seen during training (SD~2.1, SDXL, Flux).

\subsection{Ablation Studies}
\label{sec:ablation}

We conduct ablation studies to determine the best configuration for DiffusionPrint. Unless otherwise stated, the default configuration uses $C=64$ output channels, no classification head, top-$k=512$ hard negatives, and geometric augmentations. In each ablation we vary one component while keeping the rest at their default values, and select the best configuration from each study before proceeding to the next. All models are trained with a batch size of $512$, a queue size of $65{,}536$ on a single NVIDIA RTX 4090.

\paragraph{Pairing strategy.}
Table~\ref{tab:ablation} compares our spatial patch pairing to a standard MoCo baseline using augmentation-only positives with the same DnCNN backbone and training protocol. Our method improves overall accuracy from 75.00\% to 86.35\% and IID accuracy from 78.00\% to 87.69\%, confirming the effectiveness of patch pairing in DiffusionPrint.

\paragraph{Number of output channels.} Table~\ref{tab:ablation} ablates the number of DnCNN output channels $C$. When $C=1$, GAP reduces to a scalar and loses all spatial feature diversity; instead, we flatten the single-channel feature map and apply a linear projector without geometric augmentations. For $C > 1$, GAP aggregates across channels and the MLP projector operates on a $C$-dimensional vector. We find that $C=64$ provides the best overall accuracy at 86.35\%, while $C=256$ offers no improvement despite higher capacity. Preserving full spatial information via flattening ($C=1$) underperforms compared with $C=64$, suggesting that entangling content and forensic signal at every spatial position makes the contrastive objective harder to optimize. This is consistent with the nature of the forensic signal we aim to capture, which is a statistical property of the generative process distributed across the entire patch rather than localized at specific pixels. At inference, the projector is discarded and the full $(B, C, H, W)$ feature map is passed to the downstream fusion framework, preserving localization through the backbone's dense spatial outputs. This follows the same pattern as MoCo and SimCLR, where GAP is used during pretraining and full spatial feature maps are used at inference.

\paragraph{Top-$k$ hard negative mining.} Table~\ref{tab:ablation} reports the effect of varying $k \in \{64, 256, 512, 1024\}$. Performance improves consistently up to $k=512$ with 86.35\% overall and 87.69\% in-domain accuracy, after which increasing to $k=1024$ provides no further benefit and slightly reduces performance, suggesting that beyond a certain number of hard negatives the optimization becomes noisier. We adopt $k=512$ as our default.

\paragraph{Classification head.} Table~\ref{tab:ablation} compares three configurations: pure contrastive training, joint training with a binary classification head (real vs. inpainted, $C_{\text{cls}}=2$), and joint training with a generator-aware classification head (real, SD~2.1, SDXL, Flux, $C_{\text{cls}}=4$). The binary head provides marginal improvement over no classification head (86.18\% vs. 86.35\% overall), suggesting that binary supervision alone does not add useful signal beyond what the contrastive objective already captures. The generator-aware variant improves overall accuracy to 88.17\% with consistent gains in-domain (89.88\%), suggesting that explicit model attribution supervision encourages the backbone to learn generator-specific fingerprints that generalize beyond the training distribution. We adopt the generator-aware classification head as our final configuration.

\begin{table}[t]
    \centering
    \caption{Ablation studies on DiffusionPrint pretraining. All: overall linear probing accuracy. IID: accuracy on in-domain  models (SD~2.1, SDXL, Flux). Bold denotes the selected configuration.}
    \label{tab:ablation}
    \begin{tabular}{llcc}
        \toprule
        & Configuration & All (\%) & IID (\%) \\
        \midrule
        \multirow{2}{*}{Pairing}       & MoCo          & 75.21 & 78.33 \\
                                       & Ours          & \textbf{86.35} & \textbf{87.69} \\

        \midrule
        \multirow{3}{*}{Channels $C$}  & 1 (flatten) & 80.36 & 81.10 \\
                                       & 64 (GAP)    & \textbf{86.35} & \textbf{87.69} \\
                                       & 256 (GAP)   & 85.00 & 85.41 \\
        \midrule
        \multirow{4}{*}{Top-$k$}       & 64   & 85.83 & 86.45 \\
                                       & 256  & 84.95 & 85.71 \\
                                       & 512  & \textbf{86.35} & \textbf{87.69} \\
                                       & 1024 & 85.52 & 85.84 \\
        \midrule
        \multirow{3}{*}{Cls head}      & None      & 86.35 & 87.69 \\
                                       & Binary    & 86.18 & 86.75 \\
                                       & Generator & \textbf{88.17} & \textbf{89.88} \\
        \bottomrule
    \end{tabular}
\end{table}

\section{Forgery Localization Results}
\label{sec:localization}

\begin{table*}[t]
    \centering
    \caption{F1 score (\%) on TGIF fully regenerated (FR) subset at threshold 0.5 for semantic (sem) and random (rand) mask subsets. NP++: Noiseprint++ modality. DP: DiffusionPrint modality trained on SD~2.1, SDXL, and Flux. DP$^\dagger$: DiffusionPrint contrastive pretraining on SD~2.1 and SDXL only; Flux variants are unseen during pretraining. Lite Baseline: simplified fusion baseline using Segformer for RGB and a CNN for the secondary signal with concatenation. All fusion frameworks are fine-tuned on semantic FR images only; random mask results reflect generalization to unseen mask types. {\small \textcolor{green!50!black}{+$\Delta$}} indicates improvement over NP++ baseline.}
    \label{tab:main-results}
    \setlength{\tabcolsep}{3pt}
    \begin{tabular}{llcccccccccc}
        \toprule
        & & \multicolumn{2}{c}{SD~2.1} & \multicolumn{2}{c}{SDXL} & \multicolumn{2}{c}{Flux schnell} & \multicolumn{2}{c}{Flux dev} & \multicolumn{2}{c}{Flux fill} \\
        \cmidrule(lr){3-4} \cmidrule(lr){5-6} \cmidrule(lr){7-8} \cmidrule(lr){9-10} \cmidrule(lr){11-12}
        Framework & Modality & sem & rand & sem & rand & sem & rand & sem & rand & sem & rand \\
        \midrule
        \multirow{3}{*}{Lite Baseline}
            & NP++         & 52\% & 23\% & 60\% & 61\% & 75\% & 72\% & 65\% & 62\% & 64\% & 59\% \\
            & DP           & 54\% {\footnotesize \textcolor{green!50!black}{+2}} & 25\% {\footnotesize \textcolor{green!50!black}{+2}} & 64\% {\footnotesize \textcolor{green!50!black}{+4}} & 68\% {\footnotesize \textcolor{green!50!black}{+7}} & 85\% {\footnotesize \textcolor{green!50!black}{+10}} & 77\% {\footnotesize \textcolor{green!50!black}{+5}} & 80\% {\footnotesize \textcolor{green!50!black}{+15}} & 73\% {\footnotesize \textcolor{green!50!black}{+11}} & 73\% {\footnotesize \textcolor{green!50!black}{+9}} & 73\% {\footnotesize \textcolor{green!50!black}{+14}} \\
            & DP$^\dagger$ & 52\% {\footnotesize \textcolor{green!50!black}{+0}} & 25\% {\footnotesize \textcolor{green!50!black}{+2}} & 65\% {\footnotesize \textcolor{green!50!black}{+5}} & 70\% {\footnotesize \textcolor{green!50!black}{+9}} & 88\% {\footnotesize \textcolor{green!50!black}{+13}} & 76\% {\footnotesize \textcolor{green!50!black}{+4}} & 81\% {\footnotesize \textcolor{green!50!black}{+16}} & 71\% {\footnotesize \textcolor{green!50!black}{+9}} & 76\% {\footnotesize \textcolor{green!50!black}{+12}} & 77\% {\footnotesize \textcolor{green!50!black}{+18}} \\
        \midrule
        \multirow{3}{*}{TruFor~\cite{guillaro2023trufor}}
            & NP++         & 50\% & 23\% & 61\% & 62\% & 80\% & 62\% & 69\% & 46\% & 65\% & 59\% \\
            & DP           & 53\% {\footnotesize \textcolor{green!50!black}{+3}} & 25\% {\footnotesize \textcolor{green!50!black}{+2}} & 67\% {\footnotesize \textcolor{green!50!black}{+6}} & 73\% {\footnotesize \textcolor{green!50!black}{+11}} & 89\% {\footnotesize \textcolor{green!50!black}{+9}} & 85\% {\footnotesize \textcolor{green!50!black}{+23}} & 83\% {\footnotesize \textcolor{green!50!black}{+14}} & 74\% {\footnotesize \textcolor{green!50!black}{+28}} & 75\% {\footnotesize \textcolor{green!50!black}{+10}} & 74\% {\footnotesize \textcolor{green!50!black}{+15}} \\
            & DP$^\dagger$ & 54\% {\footnotesize \textcolor{green!50!black}{+4}} & 23\% {\footnotesize \textcolor{green!50!black}{-}} & 67\% {\footnotesize \textcolor{green!50!black}{+6}} & 69\% {\footnotesize \textcolor{green!50!black}{+7}} & 87\% {\footnotesize \textcolor{green!50!black}{+7}} & 77\% {\footnotesize \textcolor{green!50!black}{+15}} & 81\% {\footnotesize \textcolor{green!50!black}{+12}} & 68\% {\footnotesize \textcolor{green!50!black}{+22}} & 73\% {\footnotesize \textcolor{green!50!black}{+8}} & 69\% {\footnotesize \textcolor{green!50!black}{+10}} \\
        \midrule
        \multirow{3}{*}{MMFusion~\cite{triaridis2024mmfusion}}
            & NP++         & 47\% & 17\% & 58\% & 61\% & 80\% & 58\% & 72\% & 50\% & 69\% & 67\% \\
            & DP           & 53\% {\footnotesize \textcolor{green!50!black}{+6}} & 24\% {\footnotesize \textcolor{green!50!black}{+7}} & 64\% {\footnotesize \textcolor{green!50!black}{+6}} & 69\% {\footnotesize \textcolor{green!50!black}{+8}} & 85\% {\footnotesize \textcolor{green!50!black}{+5}} & 77\% {\footnotesize \textcolor{green!50!black}{+19}} & 80\% {\footnotesize \textcolor{green!50!black}{+8}} & 73\% {\footnotesize \textcolor{green!50!black}{+23}} & 76\% {\footnotesize \textcolor{green!50!black}{+7}} & 78\% {\footnotesize \textcolor{green!50!black}{+11}} \\
            & DP$^\dagger$ & 50\% {\footnotesize \textcolor{green!50!black}{+3}} & 22\% {\footnotesize \textcolor{green!50!black}{+5}} & 62\% {\footnotesize \textcolor{green!50!black}{+4}} & 72\% {\footnotesize \textcolor{green!50!black}{+11}} & 87\% {\footnotesize \textcolor{green!50!black}{+7}} & 82\% {\footnotesize \textcolor{green!50!black}{+24}} & 82\% {\footnotesize \textcolor{green!50!black}{+10}} & 78\% {\footnotesize \textcolor{green!50!black}{+28}} & 75\% {\footnotesize \textcolor{green!50!black}{+6}} & 76\% {\footnotesize \textcolor{green!50!black}{+9}} \\
        \bottomrule
    \end{tabular}
    \vspace{2pt}
\end{table*}

We evaluate DiffusionPrint as a forensic signal within three fusion-based IFL frameworks: TruFor~\cite{guillaro2023trufor}, MMFusion~\cite{triaridis2024mmfusion}, and a lightweight baseline (Lite Baseline). In all cases, DiffusionPrint replaces Noiseprint++ as a drop-in feature extractor. We also trained other fusion methods ~\cite{karageorgiou2024fusion, liu2025mun}, but omit them due to low performance. The retained frameworks span complex cross-modal fusion (TruFor, MMFusion) and simple fusion (Lite Baseline) with varying modality combinations. 

\subsection{Experimental Setup}
\label{sec:localization-setup}

\paragraph{Lightweight fusion baseline.}
We evaluate both forensic signals within established frameworks (TruFor, MMFusion) and a simplified two-stream model (\emph{Lite Baseline}) to isolate each modality’s discriminative power. The RGB stream uses a SegFormer (MiT-B2)~\cite{xie2021segformer} encoder, while the secondary modality is processed by a lightweight CNN. We replace the cross-attention and feature rectification modules~\cite{zhang2023cmx} with simple concatenation and $1{\times}1$ convolutions at each scale, followed by a standard All-MLP decoder. With roughly half the parameters of TruFor ($\sim$36M), this baseline attributes gains to the signal rather than fusion complexity. Full details are in the supplementary material.

\paragraph{Training.}
Each fusion framework is fine-tuned exclusively on the fully regenerated (FR) training split of TGIF2~\cite{mareen2026tgif2extendedtextguidedinpainting}, using either Noiseprint++ or DiffusionPrint as the secondary modality. We monitor the TGIF validation split to select the best checkpoint for testing. During fine-tuning, the DiffusionPrint and NoisePrint++ backbone (Section~\ref{sec:pretraining}) remains frozen; only the fusion mechanisms and decoding heads are updated. For TruFor and MMFusion, we follow the original training setups, while the Lite Baseline adopts the TruFor setup.

\paragraph{Evaluation.}
We evaluate on the TGIF test split, reporting F1 at threshold 0.5 for each generative model (SD~2.1, SDXL, Flux~1 schnell, Flux~1 dev, and Flux~1 fill dev). Results are partitioned into semantic and random mask subsets; since only semantic mask splits are used for training, the random mask subset serves as a generalization test for non-object-centric masks. To assess cross-model generalization, we additionally evaluate using our DP$^\dagger$ backbone, pretrained strictly on SD~2.1 and SDXL images, ensuring all Flux variants remain unseen during pretraining and testing whether DiffusionPrint captures a fundamental generative signal rather than model-specific shortcuts.

\begin{figure*}[t]
    \centering
    \setlength{\tabcolsep}{2pt} 
    \renewcommand{\arraystretch}{0.5} 
    
    \begin{tabular}{cccccccc}
        \scriptsize Source & \scriptsize Mask & 
        \scriptsize TruFor (NP++) & \scriptsize TruFor (DP) & 
        \scriptsize MMFusion (NP++) & \scriptsize MMFusion (DP) & 
        \scriptsize Lite (NP++) & \scriptsize Lite (DP) \\
        
        \includegraphics[width=0.115\linewidth]{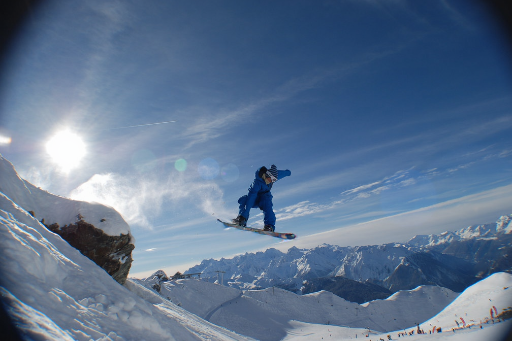} &
        \includegraphics[width=0.115\linewidth]{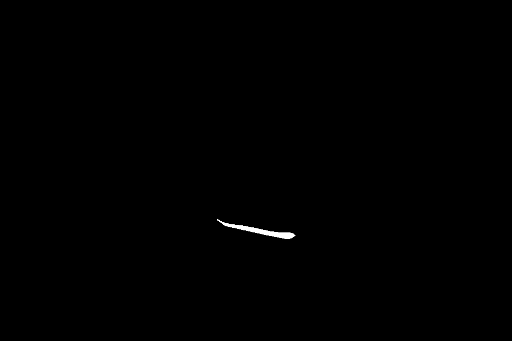} &
        \includegraphics[width=0.115\linewidth]{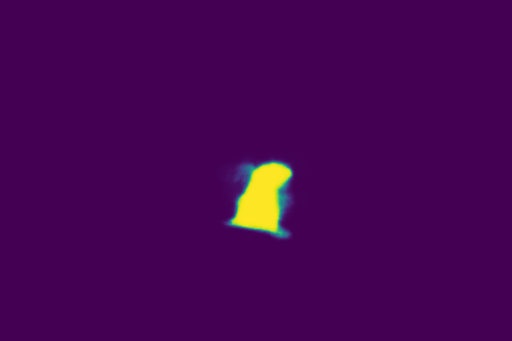} &
        \includegraphics[width=0.115\linewidth]{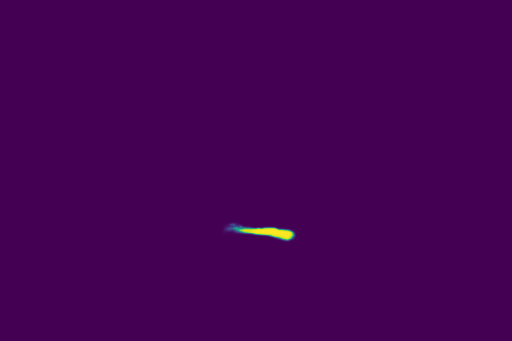} &
        \includegraphics[width=0.115\linewidth]{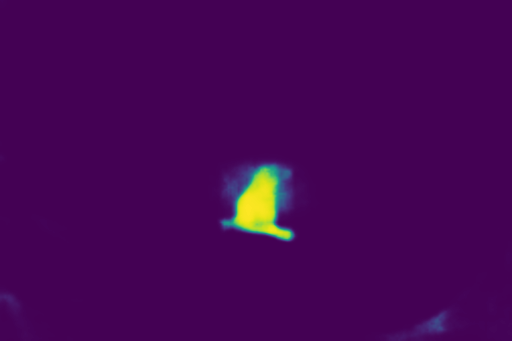} &
        \includegraphics[width=0.115\linewidth]{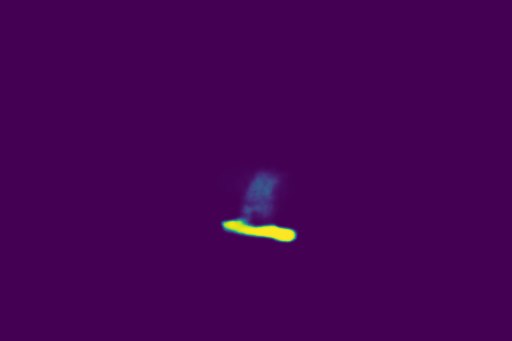} &
        \includegraphics[width=0.115\linewidth]{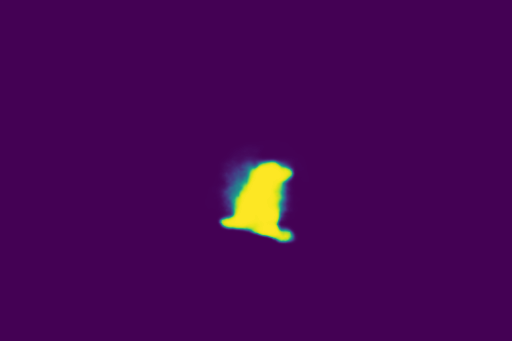} &
        \includegraphics[width=0.115\linewidth]{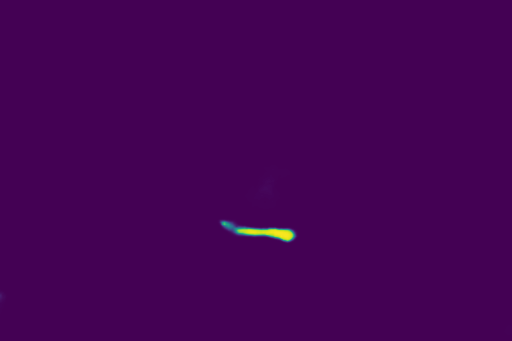} \\
        \vspace{1pt} \\ 
        
        \includegraphics[width=0.115\linewidth]{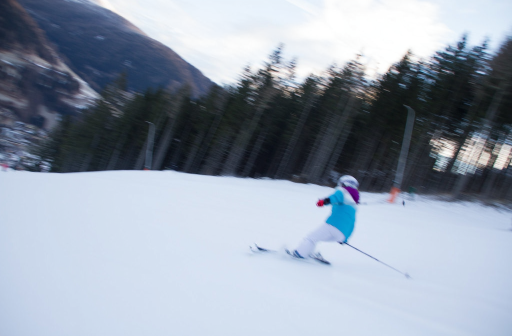} &
        \includegraphics[width=0.115\linewidth]{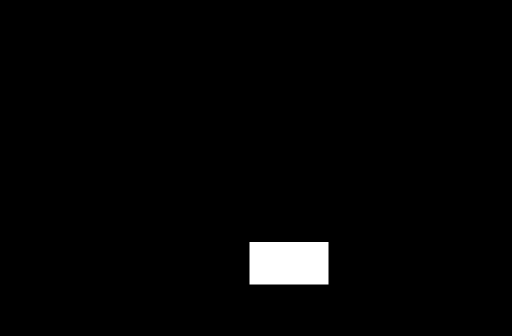} &
        \includegraphics[width=0.115\linewidth]{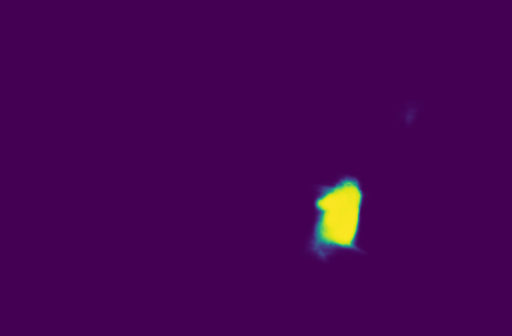} &
        \includegraphics[width=0.115\linewidth]{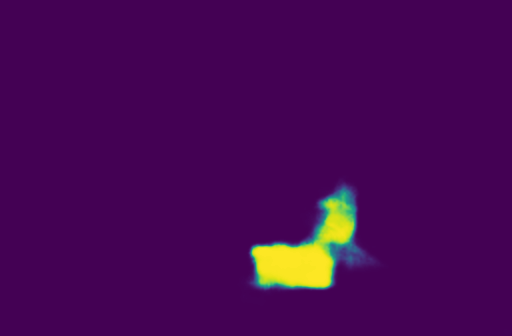} &
        \includegraphics[width=0.115\linewidth]{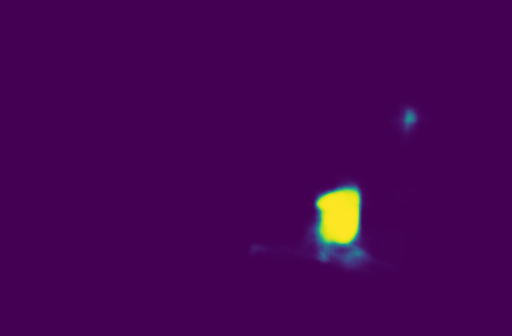} &
        \includegraphics[width=0.115\linewidth]{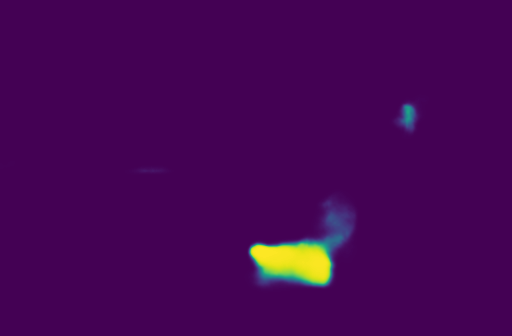} &
        \includegraphics[width=0.115\linewidth]{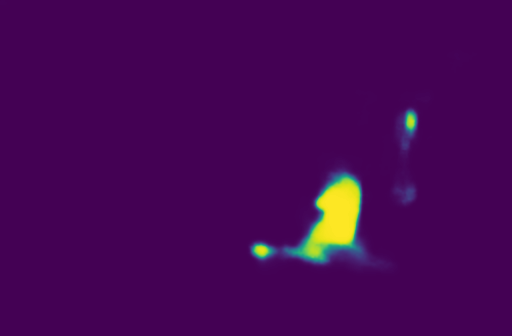} &
        \includegraphics[width=0.115\linewidth]{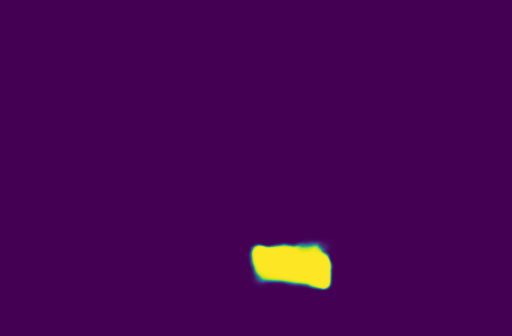} \\
        \vspace{1pt} \\
        
        \includegraphics[width=0.115\linewidth]{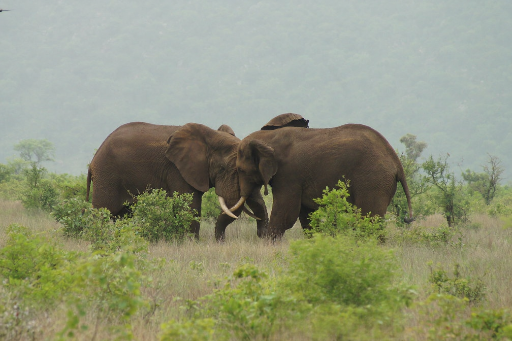} &
        \includegraphics[width=0.115\linewidth]{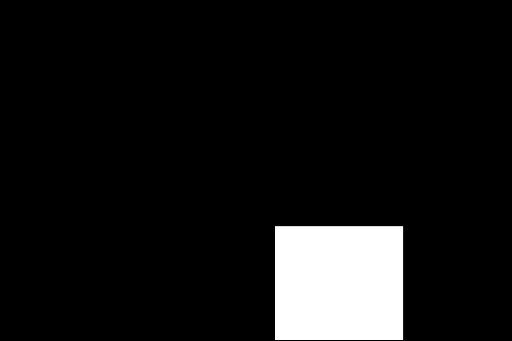} &
        \includegraphics[width=0.115\linewidth]{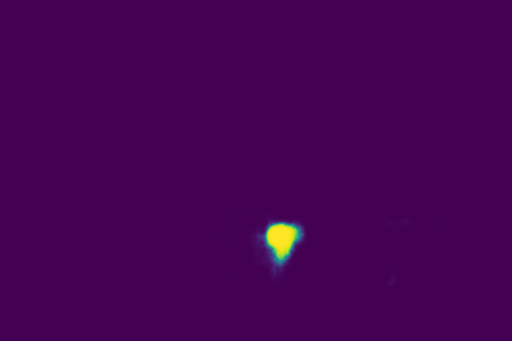} &
        \includegraphics[width=0.115\linewidth]{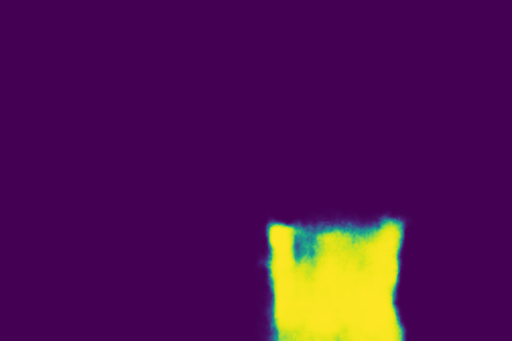} &
        \includegraphics[width=0.115\linewidth]{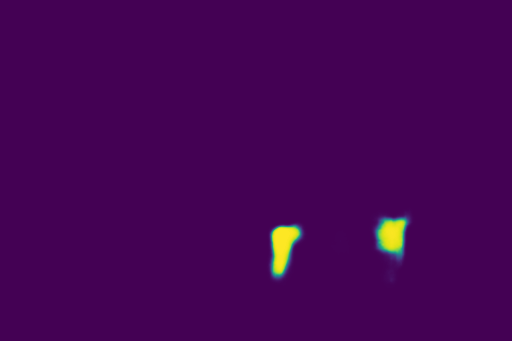} &
        \includegraphics[width=0.115\linewidth]{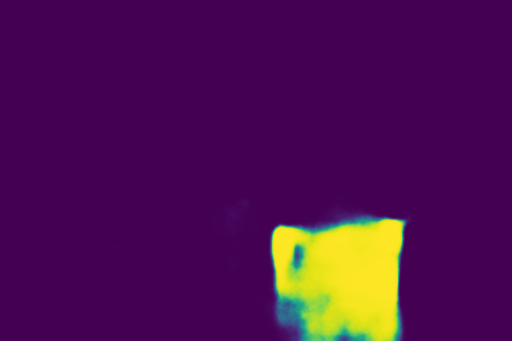} &
        \includegraphics[width=0.115\linewidth]{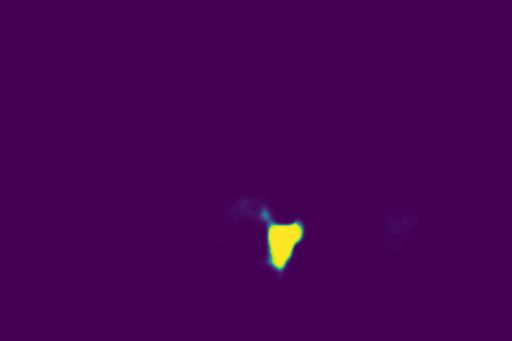} &
        \includegraphics[width=0.115\linewidth]{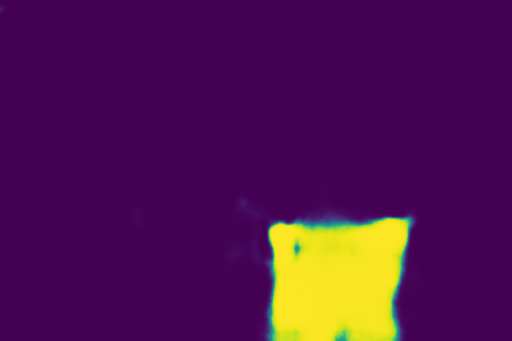} \\
        \vspace{1pt} \\
        
        \includegraphics[width=0.115\linewidth]{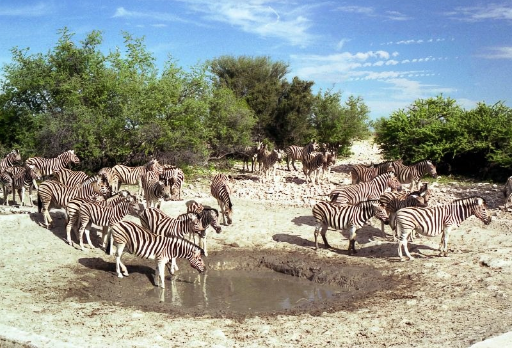} &
        \includegraphics[width=0.115\linewidth]{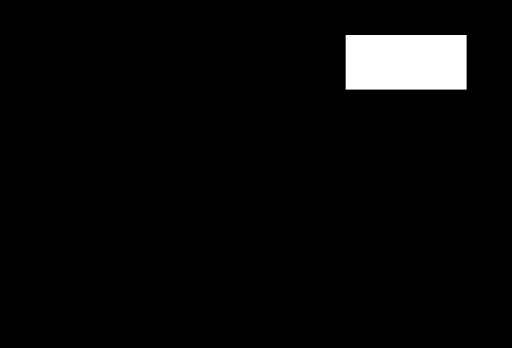} &
        \includegraphics[width=0.115\linewidth]{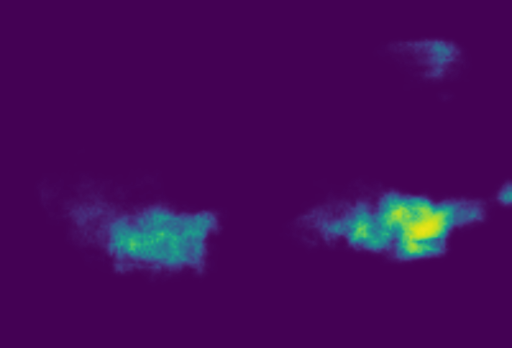} &
        \includegraphics[width=0.115\linewidth]{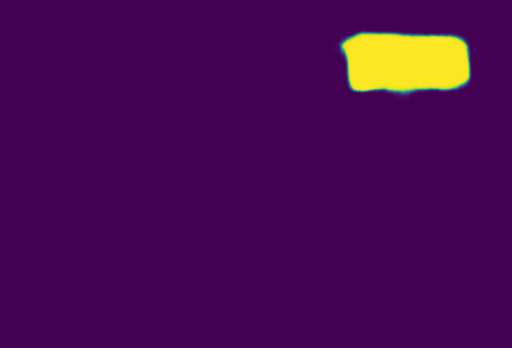} &
        \includegraphics[width=0.115\linewidth]{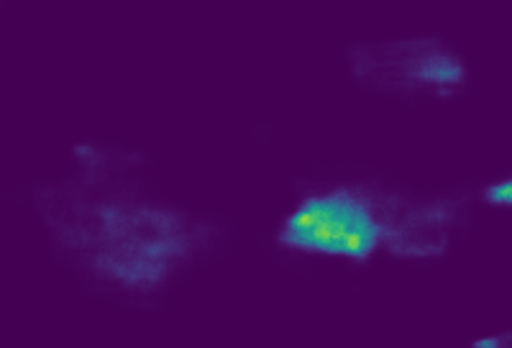} &
        \includegraphics[width=0.115\linewidth]{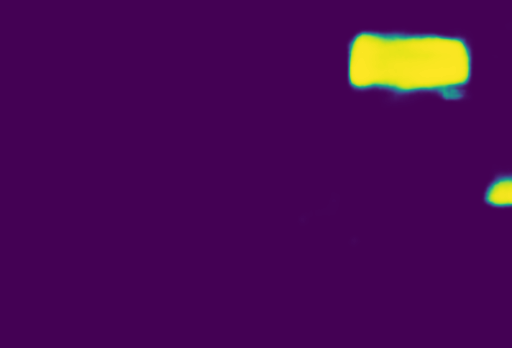} &
        \includegraphics[width=0.115\linewidth]{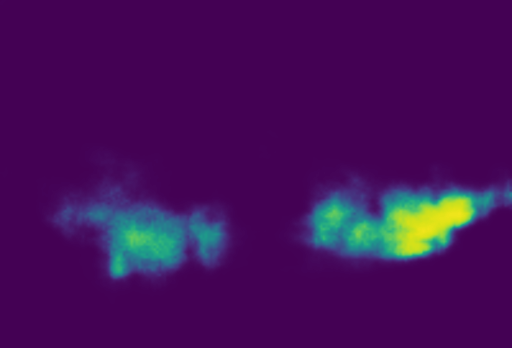} &
        \includegraphics[width=0.115\linewidth]{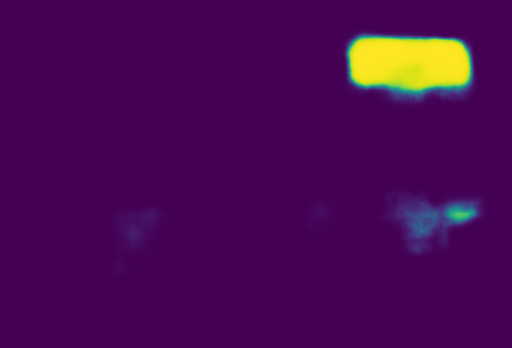} \\
        
    \end{tabular}
    
    \caption{Qualitative forgery localization results. Comparison between traditional noise-based extractors (NP++) and our proposed DiffusionPrint (DP) integrated into TruFor, MMFusion, and a Lite Baseline (Concat). DiffusionPrint consistently produces cleaner, more accurate localization maps across all frameworks.}
    \label{fig:qualitative_results}
\end{figure*}

\begin{figure}[t]
    \centering
    \begin{tabular}{cc}
        \scriptsize Input & \scriptsize Loc Map \\
        \includegraphics[width=0.46\linewidth]{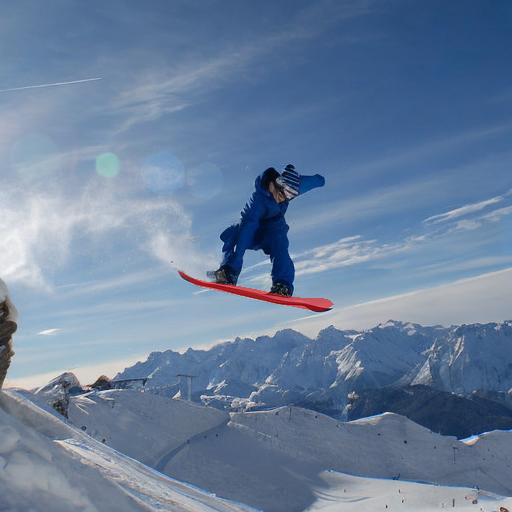} &
        \includegraphics[width=0.46\linewidth]{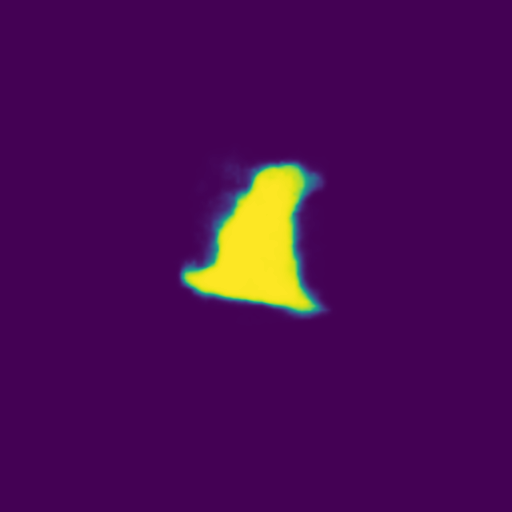} \\[4pt]
        \includegraphics[width=0.46\linewidth]{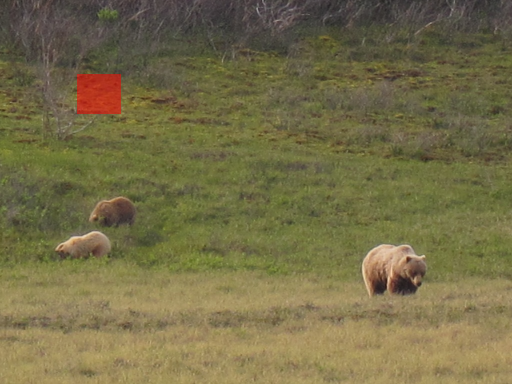} &
        \includegraphics[width=0.46\linewidth]{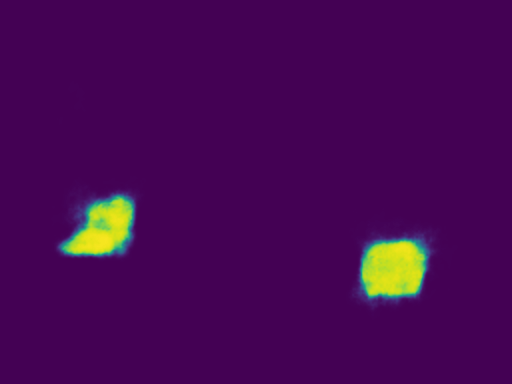} \\
    \end{tabular}
    \caption{Failure cases. \textit{Top:} an SD~2.1 inpainting correctly localized, but with an activation on an authentic object too. \textit{Bottom:} a case exhibiting object bias, where the model activates on salient non-inpainted regions rather than the manipulated area. Input images show the inpainted region outlined by the ground-truth mask.}
    \label{fig:failure}
\end{figure}

\subsection{Localization Results}
\label{sec:loc-results}
Table~\ref{tab:main-results} reports F1 scores on the TGIF FR subset for Lite Baseline, TruFor, and MMFusion using Noiseprint++ or DiffusionPrint. DiffusionPrint consistently outperforms Noiseprint++ across all frameworks and generative models. The largest gains occur on the random mask subset, a setting unseen during training, suggesting DiffusionPrint captures a more generalizable signal rather than mask-specific artifacts. Gains on semantic masks are also consistent (+2\% to +15\%). Flux variants benefit the most, with large improvements on Flux dev and Flux schnell under random masks, indicating generalization beyond U-Net-based models. In the generalization setting (DP$^\dagger$), trained only on SD~2.1 and SDXL (all Flux unseen), DiffusionPrint achieves competitive or superior performance on several Flux variants and consistently outperforms Noiseprint++, demonstrating a transferable generative fingerprint.

\subsection{Qualitative Analysis}
Figure~\ref{fig:qualitative_results} shows localization maps for four example images. Noiseprint++ often struggles under fully generative inpainting, producing noisy predictions or activating on authentic regions. In contrast, DiffusionPrint yields more precise localization with fewer false positives across all frameworks, aligning with the quantitative results and indicating a more reliable signal under latent reconstruction.

\subsection{Failure Cases}
Figure~\ref{fig:failure} shows two failure cases of TruFor with our forensic signal.
Despite overall gains, SD~2.1 inpaintings can yield spurious activations on authentic objects even when the manipulated region is correctly localized (top). The model also exhibits object bias, activating on salient non-inpainted regions rather than the manipulated area (bottom). These cases highlight that diffusion-based inpainting detection remains an open problem.
\section{Conclusion}
We presented DiffusionPrint, a patch-level contrastive framework tailored for fully-regenerative diffusion inpainting, where global latent reconstruction destroys the sensor noise used by existing IFL methods. It learns a forensic signal invariant to reconstruction-induced spectral shifts on authentic regions while capturing the generative fingerprint of inpainted content. This is achieved via a MoCo-style objective, cross-category hard negative mining, and a generator-aware classification head. Contrastive pretraining yields 88.17\% linear probing accuracy and generalizes strongly to 22 unseen generative models. As a drop-in modality for TruFor, MMFusion, and a lightweight baseline, DiffusionPrint consistently outperforms Noiseprint++, achieving up to +28\% gains on unseen mask types. The learned signal is also highly model-agnostic: training solely on SD 2.1 and SDXL generalizes competitively to unseen Flux variants.

\paragraph{Future work.}
A natural direction is integrating DiffusionPrint into a unified IFL pipeline combining generative fingerprint and camera-noise modalities, enabling robust localization across both fully-regenerative and traditional spliced manipulations without prior knowledge of the manipulation type. A systematic comparison against other self-supervised forensic signals would further clarify where the generative fingerprint provides the most benefit.

\subsubsection*{Acknowledgments} This work received funding from the Horizon Europe projects AI-CODE (grant agreement No. 101135437) and AI4TRUST (grant agreement No. 101070190).
{
    \small
    \bibliographystyle{ieeenat_fullname}
    \bibliography{main}
}

\clearpage
\setcounter{page}{1}
\maketitlesupplementary

\section{Augmentation Strategies}
\label{sec:supp_augmentations}

Contrastive learning relies heavily on data augmentation to generate diverse, positive views of the same underlying instance. However, in the context of image forensics, augmentations must be chosen carefully to avoid destroying the delicate traces left by the generative process. In this section, we evaluate the impact of different augmentation strategies during the pretraining of the DiffusionPrint backbone. Table~\ref{tab:supp_aug_ablation} reports the linear probing accuracy across four configurations: no augmentation, geometric transformations, high-pass filtering, and JPEG compression.

\begin{table}[h]
    \centering
    \caption{Ablation on pretraining augmentations. All metrics denote overall linear probing accuracy. While geometric augmentations improve representation quality, applying high-pass filters or JPEG compression severely degrades the learned forensic signal.}
    \label{tab:supp_aug_ablation}
    \begin{tabular}{lc}
        \toprule
        Augmentation Strategy & Overall Accuracy (\%) \\
        \midrule
        None & 84.90 \\
        Geometric & \textbf{86.35} \\
        Geometric + JPEG & 83.40 \\
        Geometric + HP Filter & 64.10 \\
        \bottomrule
    \end{tabular}
\end{table}

As a baseline, training the encoder with no augmentations yields an accuracy of 84.90\%. Using purely geometric augmentations (random crops and horizontal/vertical flips) provides spatial diversity and improves the accuracy to 86.35\%. These spatial transformations help the model learn a better representation without altering the local pixel statistics produced by the generative models.

In traditional image forensics, high-pass (HP) filters are often used to suppress image semantics and isolate high-frequency camera noise. However, applying an HP filter (blocking normalized frequencies $f_{\text{norm}} \leq 0.1$) during pretraining drops the accuracy to 64.10\%. This suggests that although low-frequency content contains semantic data, it still holds useful forensic information about the generative process. Filtering it out simply removes these important traces. We also tested applying random JPEG compression (Quality Factor between 70 and 95, with a 50\% probability) alongside geometric augmentations. While JPEG augmentation is commonly used in downstream tasks to improve robustness, applying it during pretraining reduces the accuracy to 83.40\%. This performance drop likely occurs because the compression step removes important forensic information that the encoder needs to effectively learn the generative fingerprint.

\section{Lite Baseline Architecture.} 
In addition to the state-of-the-art frameworks, we evaluate a custom lightweight two-stream baseline (Lite Baseline) to isolate the effectiveness of the forensic modalities with a simpler fusion mechanism. The RGB stream utilizes an ImageNet-pretrained Mix Transformer encoder (MiT-B2) from the SegFormer architecture~\cite{xie2021segformer}. For the forensic stream, the extracted feature map (either Noiseprint++ or our frozen DiffusionPrint) is passed through a lightweight convolutional secondary encoder utilizing residual blocks to produce feature maps at four corresponding scales (64, 128, 320, and 512 channels). 

Unlike the complex cross-attention mechanisms from the cmx architecture \cite{zhang2023cmx} used in TruFor and MMFusion, our Lite Baseline employs a straightforward multi-scale concatenation fusion: at each of the four scales, the RGB and forensic feature maps are concatenated along the channel dimension and reduced via a $1 \times 1$ convolution, Batch Normalization, and a ReLU activation. The fused multi-scale features are then passed into a standard SegFormer All-MLP decoder to produce the final localization map. This streamlined architecture significantly reduces the computational overhead, containing approximately 36M parameters—nearly half the size of the TruFor network.

\section{Training Details}
\label{sec:supp_setup}

For the integration of DiffusionPrint into the TruFor~\cite{guillaro2023trufor} and MMFusion~\cite{triaridis2024mmfusion} frameworks, we retain their original architectural designs and training protocols, referring readers to the respective papers for exhaustive network details. The lite baseline was adapted from the TruFor implementation. Across all frameworks, input images are randomly cropped to $512 \times 512$ and trained for 100 epochs using an SGD optimizer with an initial learning rate of 0.005 and a momentum of 0.9. To match TruFor's original effective batch size of 18, we utilize a physical batch size of 9 coupled with 2 gradient accumulation steps. Similarly, for MMFusion, we maintain the original effective batch size of 24 by employing a physical batch size of 8 with 3 gradient accumulation steps. All framework-specific hyperparameters remain strictly as originally proposed. Prior to extracting the forensic feature maps, we apply identical data augmentations to the RGB inputs across all models: random resizing in the range $[0.5, 1.5]$ and JPEG compression with a quality factor uniformly sampled between 30 and 100.

\end{document}